\documentclass[letterpaper, 10 pt, conference]{ieeeconf}  

\IEEEoverridecommandlockouts                              

\usepackage{graphics} 
\usepackage{amsmath} 
\usepackage{amssymb}  
\usepackage{graphicx}
\usepackage{subfig}

\usepackage{caption}
\usepackage[dvipsnames]{xcolor}
\usepackage[linktocpage,pagebackref=true]{hyperref}
\hypersetup{
    colorlinks=true,
    linkcolor=RoyalBlue,
    filecolor=magenta,      
    urlcolor=MidnightBlue,
    citecolor=green,
}

\title{\LARGE \bf
Efficient and Accurate Candidate Generation \\for Grasp Pose Detection in SE(3)}

\author{Andreas ten Pas$^{1}$, Colin Keil$^{1}$, and Robert Platt$^{1}$
\thanks{$^{1}$All authors are with Khoury College of Computer Sciences, Northeastern University, Boston, MA, USA
        {\tt\small atp at ccs neu edu}}%
}

\begin{document}

\bstctlcite{IEEEexample:BSTcontrol} 

\maketitle
\thispagestyle{empty}
\pagestyle{empty}

\begin{abstract}

Grasp detection of novel objects in unstructured environments is a key
capability in robotic manipulation. For 2D grasp detection problems where
grasps are assumed to lie in the plane, it is common to design a fully
convolutional neural network that predicts grasps over an entire image in one
step. However, this is not possible for grasp pose detection where grasp poses
are assumed to exist in \textsc{SE(3)}. In this case, it is common to approach
the problem in two steps: grasp candidate generation and candidate
classification~\cite{tenpas_ijrr2017,gualtieri_corl2018,mousavian_iccv2019,liang_icra2019}.
Since grasp candidate classification is typically expensive, the problem
becomes one of efficiently identifying high quality candidate grasps. This
paper proposes a new grasp candidate generation method that significantly
outperforms major 3D grasp detection baselines. Supplementary material is
available at \href{https://atenpas.github.io/psn/}{this website}.

\end{abstract}

\section{INTRODUCTION}
For many robot manipulation tasks, grasping is a key step. In order to grasp an object successfully, the robot must place its hand in exactly the right position and orientation before attempting to close its fingers. Grasp success depends on the geometry, friction, mass, and deformability of both the robot hand and the object. Possible collisions with nearby objects further complicate the grasping process. 

There have been several recent approaches that employ deep learning techniques
to predict grasp poses given point cloud data as input~\cite{lenz_ijrr2015,
mahler_rss2017, kalashnikov_corl2018, levine_ijrr2018}. However, while these
methods have been shown to enable a robot to grasp a large variety of rigid
objects, with many different shapes, sizes, and textures, they are often
restricted to grasp detection in the plane or with a shallow out-of-plane
orientation. This is a severe limitation because it prohibits the object from
being grasped from arbitrary directions in \textsc{SE(3)} and thus restricts
how the object can be manipulated.

Some recent work has focused on grasp detection in SE(3) that can exploit all the
freedoms of positioning and orienting a robotic
hand~\cite{tenpas_ijrr2017,gualtieri_corl2018,mousavian_iccv2019,qin_corl2019}. Such methods often follow a \textit{generate-and-test} strategy, where a number of grasp candidate poses are first generated and then tested for a grasp quality criterion using a learned classifier. The key question here is how to generate grasp proposals. Many methods~\cite{tenpas_ijrr2017, tenpas_isrr2015, liang_icra2019} either generate large numbers of grasp proposals exhaustively or use heuristics to prune geometrically infeasible proposals. These methods can be slow because they typically perform collision checks and/or grasp stability tests against the observed scene geometry~\cite{tenpas_ijrr2017,qin_corl2019}. An important exception to the above is the recent work of Mousavian~\cite{mousavian_iccv2019} who uses the decoder portion of a variational autoencoder to propose grasp poses. This is a more flexible approach, but it can be slow.

In this paper, we ask whether it is possible to match or exceed the performance
of existing baselines such as~\cite{tenpas_ijrr2017}
and~\cite{mousavian_iccv2019} using a relatively simple neural network model
architecture to generate grasp proposals from 3D point clouds. Our method first
samples a set of points from the cloud. For each sample, it efficiently encodes
the geometry of a region about the sample by cropping local rectangular patches
from each of several orthographic projections of the global scene. This
representation of the local region is input to a neural network model that
predicts which orientations about the sampled point are likely to contain a
grasp. This method is computationally more efficient than prior
methods~\cite{tenpas_ijrr2017,mousavian_iccv2019} because both the cropping
process and the candidate generation process are very quick. We demonstrate in
simulation that both the accuracy and speed of this method compare favorably
with previous approaches and we show that the method is effective in practice
on a real robot.


\section{Related Work}

\subsection{Object Proposals in Computer Vision}
The ideas for proposing grasp poses algorithmically are analogous to proposing
object locations and bounding boxes in computer vision. Girshick et al. laid the
groundwork~\cite{girshick_cvpr2014} in which a number of candidate object
regions are proposed and individually evaluated by a convolutional neural
network (CNN). Their work was later extended with region of interest pooling
which improves object detection accuracy and reduces computational
cost~\cite{girshick_iccv2015}. Ren et al.~\cite{shaoqing_nips2015} further
improved object detection by predicting candidate object bounding boxes with a
Region Proposal Network (RPN), extracting features for each box with region of
interest pooling, classifying the box content and regressing the box
parameters. Features for the RPN and the box evaluation are
shared to further reduce computational inference cost. By learning to predict
offsets from fixed reference bounding boxes, called \textit{anchors}, the
authors predict bounding boxes of different aspect ratios and sizes.
Contrastively, Liu et al.~\cite{liu_eccv2016} used a single network that makes
predictions about a discretized space of default bounding boxes whose aspect
ratios and scales are fixed. Similarly, our approach learns to predict 3D
rotational ``offsets'' from a fixed set of reference grasp poses.

\subsection{6-DOF Grasp Pose Detection}
The state-of-the-art in grasp pose detection are deep learning based methods.
Much work has focused on 3- or 4-DOF grasp configurations~\cite{lenz_ijrr2015,
mahler_rss2017, kalashnikov_corl2018, levine_ijrr2018, satish_ram2019,
bodnar_rss2020}. Typically, such approaches keep two orientation DOFs fixed and
only learn the orientation for one axis, usually the grasp approach axis.  This
restricts subsequent manipulation tasks because it severely limits the ways in
which an object can be grasped.

Another avenue of recent research explored 6-DOF grasping. Mousavian et al.
used a variational auto-encoder to sample a diverse set of grasp poses for an
object~\cite{mousavian_iccv2019}. These poses are then iteratively evaluated
with a separate network and refined using the evaluator network's gradient. All
of their networks are based on the PointNet++ architecture~\cite{qi_nips0217}.
This approach was extended by Murali et al.~\cite{murali_icra2020} to grasping
a particular object in clutter. Qin et al.~\cite{qin_corl2019} used a
single-shot grasp proposal network that predicts grasp poses by regression
directly from the complete visible scene.  Their network is based on the
PointNet++ architecture~\cite{qi_nips0217}. In prior work, we detected grasp
poses by first generating grasp proposals based on local geometry and then
classifying multi-view projected features for each of the
proposals~\cite{tenpas_ijrr2017}. Liang et al. extended this approach to
directly work with point cloud data by training a network based on the
PointNet~\cite{liang_icra2019} architecture. Gualtieri and Platt learn to grasp
as part of pick and place modeled as an MDP with abstract state and action
representations, and solve this MDP with a hierarchical method that tells the
robot where to look in the scene~\cite{gualtieri_corl2018}. Zhou and
Houser~\cite{zhou_rss2017} learn to predict a grasp score from a depth image
and use the score for grasp pose refinement. Yan et al.~\cite{yan_icra2018} use
generative 3D shape modeling to learn scene reconstruction that is then used to
predict grasp outcomes. Merwe et al.~\cite{merwe_icra2020} learn to reconstruct
the object geometry and predict a score for a multi-fingered robot hand
configuration that consists of a 6-DOF hand pose and finger joint positions. Lu
et al.~\cite{lu_ram2020} train a 3D CNN on voxels to predict grasp success for
a 16-DOF hand. Wu et al.~\cite{wu_ar2020} train a network to iteratively decide
whether to zoom in or to execute a grasp for multi-fingered hands. Varley et
al.~\cite{varley_iros2017} train a 3D convolutional network to perform object
shape completion and then find grasps for the completed object.  In contrast to
the literature, we directly predict grasp poses based on the point cloud
geometry and do not try to complete objects. Instead of generating grasp
proposals in a heuristic fashion, we use a neural network to evaluate a fixed
set of grasp poses.


\section{Problem Definition}

We define a robot, $\mathcal{R}$, as a robot hand that can move to an arbitrary pose in its workspace. The \textbf{robot state} is defined by the pose of the hand, $h \in SE(3)$, and the configuration of the fingers. In this work, we assume the hand is a parallel jaw gripper whose jaws are actuated by a single degree of freedom. The \textbf{world state} $\mathcal{W} \in \mathbb{W}$ is a
description of the state of the environment and objects around the robot, where
$\mathbb{W}$ is the set of all possible world states. A \textbf{point cloud},
$\mathcal{C} \in \mathbb{C}$, is a finite set of points in the robot's
environment which are obtained by one or more depth sensors, where $\mathbb{C}$ denotes the space of possible point clouds that can
be generated by the sensor arrangement of the robot. The world state is a
latent variable that is observed only via the point cloud. In particular, we
model the depth sensor(s) as a function $\Lambda : \mathbb{W} \rightarrow
\mathbb{C}$ that encodes aspects of world state as a point cloud. Given a robot
$\mathcal{R}$ and a point cloud $\mathcal{C} = \Lambda(\mathcal{W})$ for some
hidden world state $\mathcal{W}$, the problem of \textbf{grasp pose detection}
is to find a robot hand pose, $h \in SE(3)$, such that if the robot hand is
moved to $h$ and the fingers are closed, then some object in the world
$\mathcal{W}$ can be grasped.

We approach grasp pose detection as a two step process. First, we
generate a set of grasp pose proposals. Second, we evaluate the probability
that each proposal is an successful grasp. In this paper, we are mainly concerned
with improving the first step. We formulate the problem of generating grasp
proposals as learning a function, $f: \mathcal{C} \rightarrow \text{SE}(3)$,
that maps from the space of possible point clouds to the space of possible
robot hand poses.


\section{Grasp Detection System}

\begin{figure}[b]
  \begin{center}
  \includegraphics[width=0.48\textwidth]{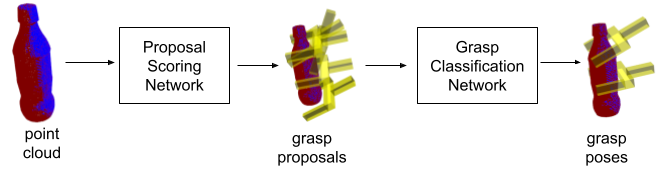}
  \end{center}
  \caption{Overview of our approach (blue: visible point cloud, red: hidden object model). Given a point cloud, we generate grasp pose proposals using a convolutional neural network. We then classify the proposals as actual grasps using another convolutional neural network.}
  \label{fig:system_diagram}
\end{figure}

Figure~\ref{fig:system_diagram} shows an overview of the end-to-end grasp detection system. There are two components: the proposal scoring network that generates a large set of grasp candidates quickly and the grasp classification network that makes a high quality binary prediction about each grasp candidate. Since the creation of images for the grasp classification network is computationally expensive~\cite{tenpas_ijrr2017}, the proposal scoring network is essential because it focuses attention on promising grasp candidates. 

\subsection{Grasp Candidate Generation}

If we were doing grasp detection in the plane (i.e. considering only $x,y,\theta$ grasp poses), then the simplest approach to grasp candidate generation would be a single forward pass through a fully convolutional neural network, e.g., as in Mahler et al.~\cite{mahler_rss2017}. However, we cannot use this approach here because we are doing grasp detection in SE(3). Instead, we do the following. We sample $n$ points from a region of interest in the point cloud. For each of the $n$ points, we  perform a single forward pass through the proposal scoring network predicting which hand orientations about the sampled point are likely to be a good grasp. 

\subsubsection{Input to the proposal scoring network}

\begin{figure}
  \begin{center}
    \includegraphics[width=0.45\textwidth]{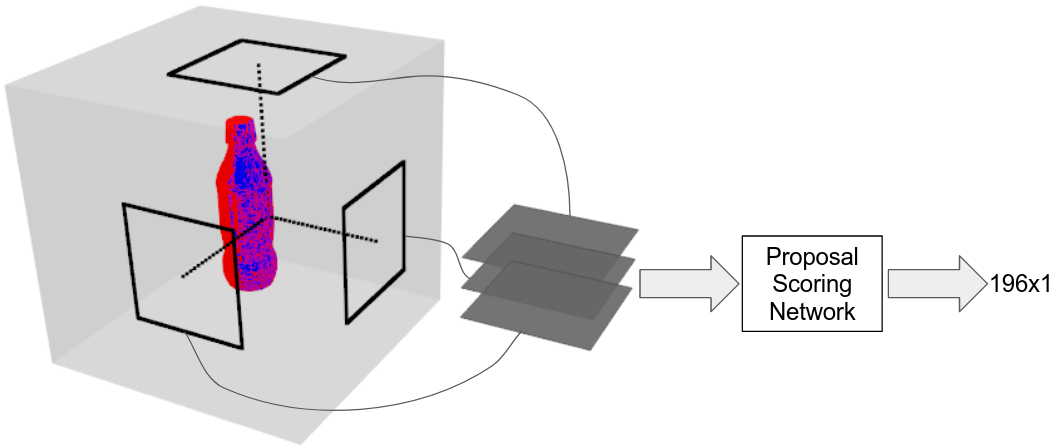}
  \end{center}
  \caption{Grasp proposals are predicted with a proposal scoring network that takes a 3-channel image and produces a score for each grasp pose represented by the image. Each channel is an orthographic representation of a subset of the point cloud.}
  \label{fig:proposal_descriptor}
\end{figure}

The proposal scoring network takes as input information about the local point cloud near the sampled point by extracting a bounding cube centered on the sampled point. Since we plan to do one forward pass through the proposal scoring network per point sampled ($k$ forward passes), it is critical that this bounding cube is encoded efficiently. We do the following. First, we create a three-view orthographic representation of the entire (partial view) point cloud expressed in the reference frame of the camera and centered on a point of interest (e.g., the estimated object center). Each of the three orthographic views is associated with a height map that describes the scene when viewed from that direction. These three orthographic projections are illustrated in Figure~\ref{fig:proposal_descriptor} as the sides to a cube that encloses the point cloud. Second, we crop a fixed-size rectangle in each of the three orthographic projections centered on the sampled point. Each of these crops encodes information about the local neighborhood of the sample (the three orthogonal boxes outlined in black in Figure~\ref{fig:proposal_descriptor}). Finally, the height maps contained in these three rectangles are stacked and input to the grasp proposal network as a three-channel image. Notice that the approach above is very efficient per sampled point. For each point, we simply crop a rectangle from each of three images and copy them into the network input as a three-channel image (one channel for each crop).

\subsubsection{Output of the proposal scoring network}

\begin{figure}
  \begin{center}
  \includegraphics[width=0.29\textwidth]{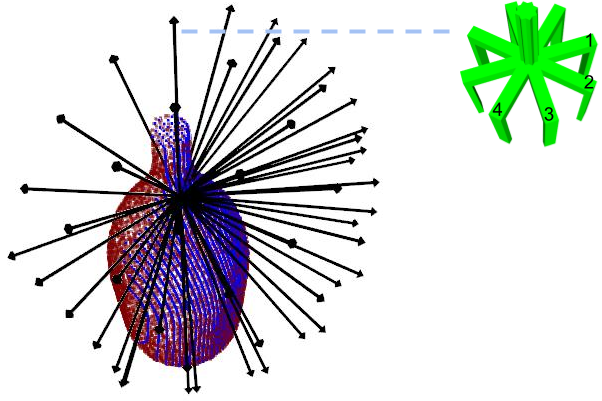}
  \end{center}
  \caption{The robot hand orientations considered by our method. Each black arrow
  corresponds to a hand approach direction. For each direction, there are
  four possible rotations about the approach axis (green).}
  \label{fig:hand_orientations}
\end{figure}

Our baseline architecture for the proposal scoring network consists of two convolutional layers (no zero padding, stride 1, kernel size 5) followed by two FC layers followed by a sigmoid layer with $m$ outputs. Each output is interpreted as the probability that a grasp exists at a particular orientation when the closing region of the hand is centered on the sampled point and the hand is ``pushed'' forward until some part of it contacts the point cloud. In this paper, we focus on the case where there are $m=196$ orientations (49 approach axes $\times$ 4 rotations about each axis), but the method should generalize to other values of $m$. Figure~\ref{fig:hand_orientations} illustrates the $m$ orientations about which our network makes predictions. These orientations cover roughly a half dome of orientations pointed in the direction of the camera and are expressed in the reference frame of the camera. Note that since the output layer is a sigmoid instead of a softmax, the network makes predictions about grasps at multiple orientations for each sampled point. To summarize, the proposal scoring network takes as input the visible point cloud geometry in the vicinity of the sample point and outputs predictions about which orientations around the sampled point are likely to be grasps.

\subsubsection{Loss Function}

Learning the parameters of the proposal scoring network is a multi-label classification problem where the labels are multi-hot vectors, $y \in [0,1]^m$, where $m$ is the number of orientations considered by the proposal scoring network. Generally, given some object part, multiple robot hand orientations can usually lead to a successful grasp. We treat this problem as if there are $m$ independent binary classification problems, i.e., the binary cross entropy loss function is calculated separately for each orientation.

\subsection{Grasp Classification}

To classify grasp proposals, we use a grasp classification network that is similar to the one presented in our earlier work~\cite{tenpas_ijrr2017}. The network takes as input information about the local point cloud that would be contained in the closing region of the robot hand at the grasp pose (one 3D rotation is required for each pose). The points in that region are orthographically projected onto a plane parallel to the hand's approach axis. A height map of these points makes up the first channel and the average surface normal at each of these points makes up three more channels of a four channel image. We use this image type as it is much faster to compute than ones with a larger number of channels~\cite{tenpas_ijrr2017}. The output of the network is a binary label that is one if the grasp is predicted to be successful and zero otherwise. The architecture for the network is the same as for the proposal scoring network. We use the cross entropy loss to train the network.


\section{Network Training}
\label{sec:network_training}
We implemented our networks in PyTorch 1.4~\cite{pazke_neurips2019}. For point cloud processing, we used Open3D 0.9~\cite{zhou_arxiv2018}. To generate synthetic scenes, we used Pyrender~\cite{matl_github2019}.

\subsection{Ground Truth Grasps}
\label{sec:ground_truth_grasps}
We consider a grasp to be successful if it (1) is collision-free and (2) has force closure. We assume a parallel jaw gripper with soft contacts so that an antipodal grasp is a sufficient and necessary condition for force closure~\cite{nguyen_ijrr1988}. We consider all points within a fixed distance from each finger as possible contact points. We then check if the surface normals of those points lie within the friction cone with a fixed friction coefficient. This is the same procedure as in our earlier work~\cite{tenpas_ijrr2017}.

\subsection{Generating Data for Training}
\label{sec:training_data}
\begin{figure}
  \begin{center}
  \includegraphics[trim=1cm 1cm 1cm 2cm, clip, width=0.38\textwidth]{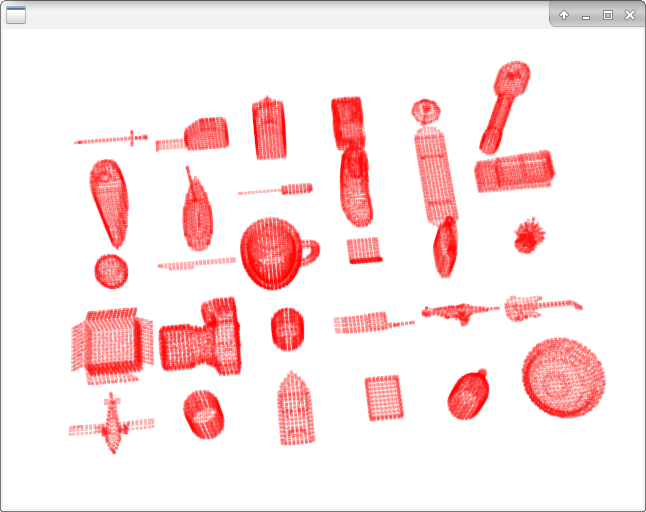}
  \end{center}
  \caption{Example instances from 3DNet object categories used for training.}
  \label{fig:3dnet_objects}
\end{figure}

Our training set consists of a total of 300 objects, ten objects for each of thirty categories in 3DNet~\cite{wohlkinger_icra2012} (see Figure~\ref{fig:3dnet_objects}). To fix holes and other issues with the 3DNet object meshes, we apply a preprocessing method that produces watertight meshes with a 2-manifold topology and vertices distributed about uniformly on the object surface~\cite{huang_arxiv2018}. We obtain point clouds from a mesh by placing it at the origin of a fixed world frame and scaling it by a randomly chosen factor such that the object's extents are within 
$[0.01\text{m}, 0.07m]$ We then obtain single-view, partial point clouds by sampling camera locations uniformly on a sphere that has a radius of 0.5m and that is centered on the world frame's origin. For our training set, we sampled 20 camera locations. From each of the resulting point clouds, we uniformly sampled 100 points and evaluated grasps with each method used for the simulation experiments against the corresponding ground truth mesh. In total, we evaluated about 117 million grasp poses to train the proposal scoring network and 600,000 poses for the grasp classification network.

\subsection{Training the Networks}

We use a batch size of 64 and images of width 60 and height 60. All models are
trained with stochastic gradient descent with a momentum of 0.9. The learning
rate starts at 0.01 and is exponentially decreased after each epoch by a factor
of 0.96. We only use the synthetic data (described above) for training.


\section{Antipodal Detection Experiments}

\label{sec:simulation_experiments}

In this section, we evaluate how well our method can detect antipodal (i.e. two finger force closure) grasps for objects presented in isolation. We obtain synthetic point clouds in the same way as described in Section~\ref{sec:network_training}.

\subsection{Object Test Set}
\label{sec:test_set}
We evaluate on novel object instances from the same thirty object categories in 3DNet which were used for training. Objects are randomly scaled and viewpoints are generated in the same way as for the training set (see Section~\ref{sec:training_data}). For each viewpoint, we sample 100 points uniformly from the cloud.

\subsection{Comparisons}

We compare our method, called QD, against two ablations, QD:GC and QD:ROT, and one baseline GPD. QD:GC is an ablation of QD that consists of only the grasp classification network (second half of Figure~\ref{fig:system_diagram}). QD:ROT is an ablation that consists of only the proposal scoring network (first half of Figure~\ref{fig:system_diagram}). Both of these ablations are one-stage grasp detectors which evaluate all potential grasp poses. The GPD baseline is the 2-stage grasp detector from~\cite{tenpas_ijrr2017}. It is structurally similar to the method in this paper, but uses a geometric proposal strategy instead of a learned proposal generator.

\subsection{Evaluation Metrics}

We evaluate our approach with three metrics: precision, recall, and detections per second (DPS). Precision and recall have the standard definitions. DPS is the number of grasp predictions produced by our method (both stages of Figure~\ref{fig:system_diagram}) per second. In addition to the standard precision/recall plot (Figure~\ref{fig:3dnet_4channels_pr}) we also show precision/DPS (Figure~\ref{fig:3dnet_4channels_pdps}). Precision/DPS shows the tradeoff between precision and the number of predictions the grasp detection system makes. Ideally, our system would achieve a point in the upper right corner of this plot -- lots of high precision grasps produced per second.

\subsection{Results: Isolated Objects}

\begin{figure}
  \begin{center}
    \includegraphics[trim=0 0 0 1.0cm, clip, width=0.48\textwidth]{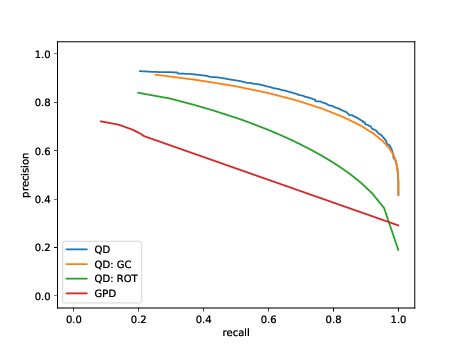}
  \end{center}
  \caption{Precision vs recall comparison on novel 3DNet object instances.}
  \label{fig:3dnet_4channels_pr}
\end{figure}

\begin{figure}
  \begin{center}
    \includegraphics[trim=0 0 0 1.0cm, clip, width=0.48\textwidth]{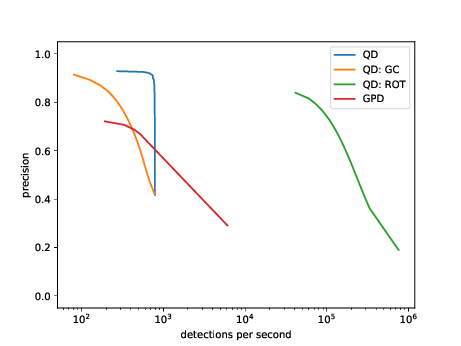}
  \end{center}
  \caption{Precision vs detections per second on novel 3DNet object instances.}
  \label{fig:3dnet_4channels_pdps}
\end{figure}

Figures~\ref{fig:3dnet_4channels_pr} and~\ref{fig:3dnet_4channels_pdps} compare QD against QD:GC, QD:ROT, and GPD for the objects in the \textit{3DNet} test set in terms of precision/recall and precision/DPS. First, notice that QD outperforms the GPD baseline by a large margin in both plots. While the QD:GC ablation is very similar to QD in terms of precision/recall, notice that it is much slower than QD, especially at high precision values. (At 90\% precision and above, QD:GC is approximately an order of magnitude slower than QD.) We attribute this speedup to the fact that our candidate generation strategy allows us to ignore a large number of candidates that the GC network would otherwise need to consider. Whereas QD:GC has to exhaustively calculate grasp images for every possible grasp proposal, QD only considers a subset. Looking at QD:ROT, notice that it is much faster than all other methods, but its precision/recall curve is significantly lower than both QD:GC and QD. This confirms that the proposal scoring network is fast but that it is not as accurate as the grasp classifier. These results emphasize that both components of QD are important for its performance: ROT allows us to subsample the set of grasp proposals to reduce runtime and GC allows us to make more accurate predictions because of the more informative grasp descriptor representation.

\begin{figure}
  \begin{center}
    \includegraphics[width=0.9\linewidth]{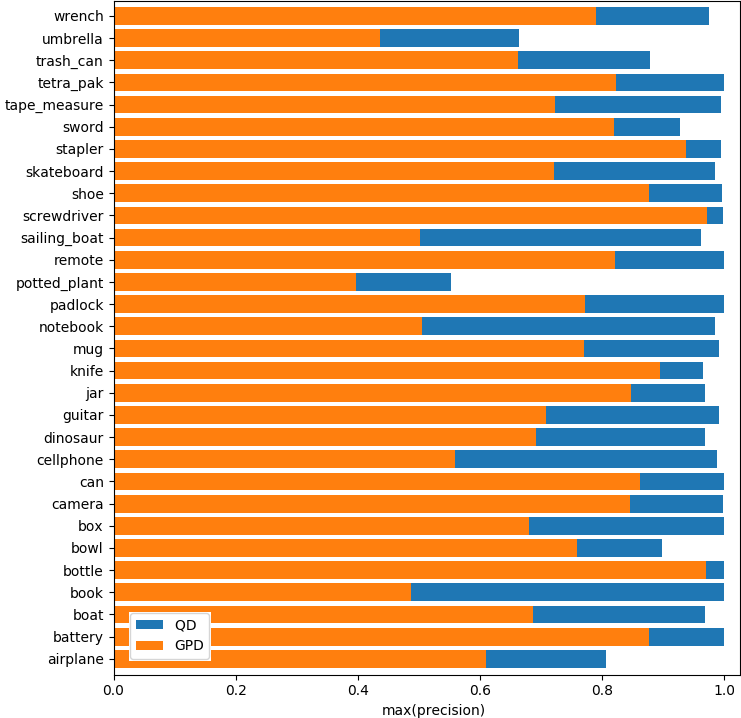}
  \end{center}
  \caption{Per-category maximum precision for GPD and QD on novel, isolated
  object instances from 3DNet.}
  \label{fig:per_category_precision}
\end{figure}

Figure~\ref{fig:per_category_precision} shows the precision that can be achieved by QD and GPD for each of the 30 object categories considered in our experiments. Notice that while QD outperforms GPD for all categories, the amount of outperformance is significant for some categories, like books. We believe that this simply reflects the difficulty the geometry based GPD proposal generator has for certain objects.

\section{PyBullet Simulation Experiments}

\label{sec:physics_simulation}

In this section, we investigate the performance of our method in a PyBullet physics simulation of robot grasping. While point clouds are still obtained synthetically in the same way as before, we measure the actual performance of grasps in the simulator.

\subsection{Simulating Grasps}
Each grasp is simulated with a Robotiq 2F-85 parallel jaw gripper, in a top down configuration. Objects are initialized at an inverted grasp pose relative to the gripper, with gravity turned off. The grasp is then tested by closing the fingers, turning on gravity, executing a shaking motion (approximately 8 seconds in simulator time), and waiting for a short time before evaluating (0.5 seconds). Any object that is not grasped or slips from the fingers quickly falls away from the gripper. In the evaluation, any object that falls below a certain distance from the gripper, or is initialized in collision, is labeled as a grasp failure. The friction coefficient between the gripper and objects is kept constant, and all objects are  simulated with a uniform mass of 0.5kg. The simulation is designed to mimic the format used in 6DOF Graspnet \cite{mousavian_iccv2019}, which is not publicly available for comparison.

To make the ground truth dataset robust against errors caused by imprecise
robot kinematics and sensor noise, we simulate a small number of randomly
perturbed poses for each grasp pose (i.e., in addition to the original
proposal). The perturbed grasp poses are generated by applying a uniformly
distributed random rotation between 0 and 5 degrees about a random axis, and a
uniformly distributed random translation between 0 and 3mm.  We decide the
final label by a majority vote over the labels of the perturbations and the
original grasp pose, i.e., if 3 out of the 5 poses are successful, we label the
grasp as positive.

\begin{figure}[b]
  \begin{center}
    \includegraphics[width=0.48\textwidth]{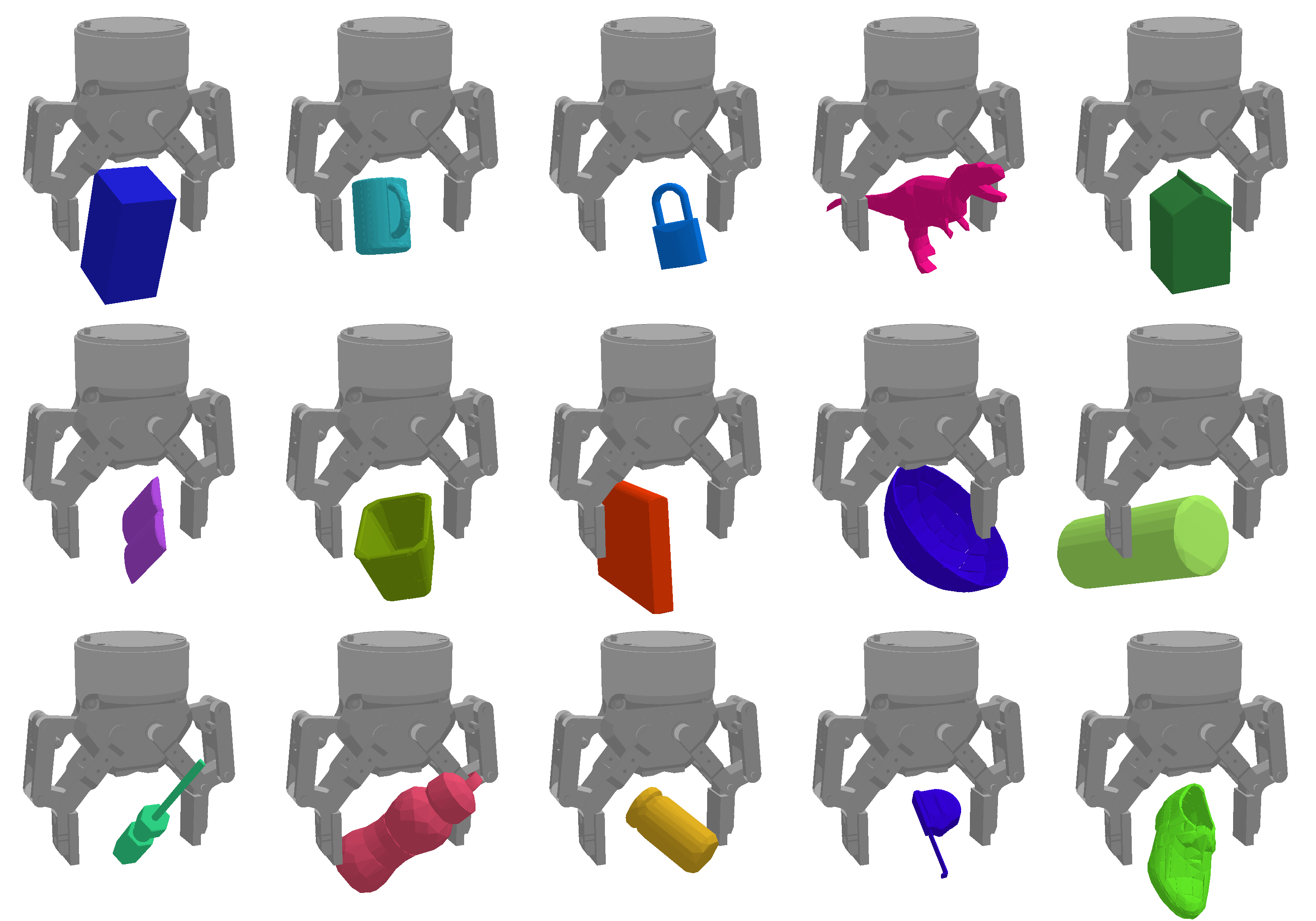}
  \end{center}
  \label{fig:sim_grasps}
  \caption{Examples of grasps simulated in PyBullet.}
  \label{fig:simulator}
\end{figure}

We compare our method to Graspnet~\cite{mousavian_iccv2019}, a recently published 6-DOF grasp pose detection method. After obtaining a point cloud using Pyrender, we run Graspnet with the following default settings: 200 samples for the generative network and 10 refinement steps using Metropolis-Hastings sampling. We then label the resulting grasp poses using the PyBullet simulator as described above. We evaluate on unseen object instances from the same object categories that Graspnet has been evaluated on: bottles, bowls, boxes, cylinders and mugs (BBBCM). Cylinders and boxes are generated using Open3d with random dimensions. Bowls, bottles, and mugs are taken from the 3DNet object set.

\subsection{Results: Physics Simulation of Grasps}

\begin{figure}[t]
  \begin{center}
  \includegraphics[trim=0 0 0 1.0cm, clip, width=0.48\textwidth]{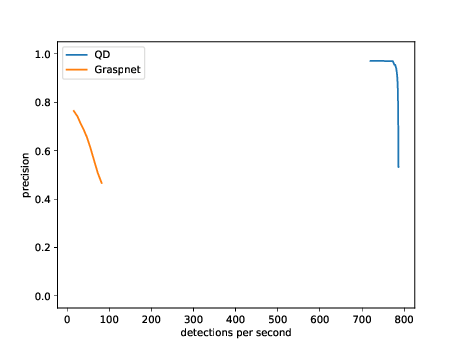}
  \end{center}
  \caption{Precision vs DPS comparison between QD and Graspnet. QD was trained using the full 3DNet object set and we use the version of Graspnet pretrained in~\cite{mousavian_iccv2019}.}
  \label{fig:pybullet_results}
\end{figure}

Figure~\ref{fig:pybullet_results} shows the comparison with Graspnet~\cite{mousavian_iccv2019}. Our method is faster and produces grasp detections with higher precision. We do not perform the precision/recall comparison here because the Graspnet proposal generator does not have a discrete hypothesis set and we therefore cannot calculate recall. However, the precision/DPS shows that our method is much faster with no worse precision. It is important to point out here that Graspnet is at a disadvantage relative to our detector. The problem is that Graspnet cannot be retrained on our training set without access to high-end GPUs. As a result, we are using the version of pretrained Graspnet weights found using the labels generated in~\cite{mousavian_iccv2019}. Although both methods were trained using the same object categories, it is likely there are slight differences between the labels in the training set and that this puts Graspnet at a disadvantage. However, given what is publicly available, this is the closest comparison that is possible.

\section{Robot Experiments}
\label{sec:robot_experiments}

\subsection{Setup}
\label{sect:robot_setup}

\begin{figure}[b]
  \begin{center}
    \subfloat[]
    {
      \includegraphics[width=0.21\textwidth]{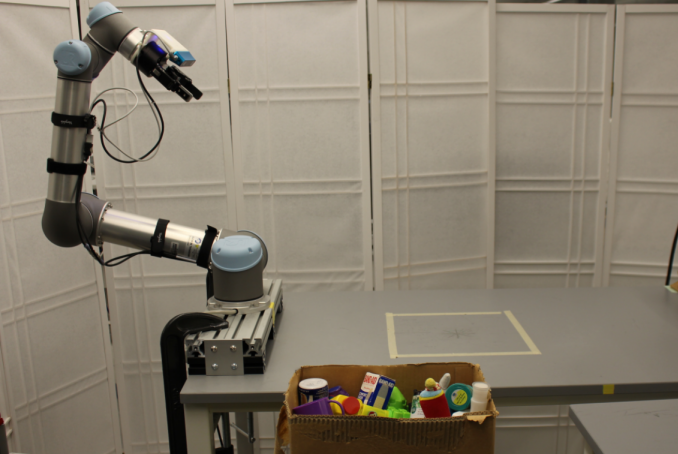}
      \label{fig:single_viewpoint}
    }
    \subfloat[]
    {
      \includegraphics[width=0.24\textwidth]{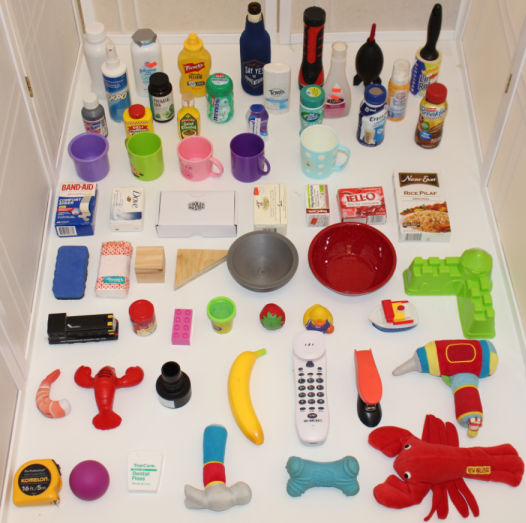}
      \label{fig:robot_objects}
    }
  \end{center}
  \caption{Robot experiments setup. (a) A point cloud is acquired from a single
  viewpoint. (b) Object set for isolated grasping.}
  \label{fig:viewpoint_and_objects}
\end{figure}

We conducted all our robot experiments on a Universal Robots' \emph{ur5} robot with six DOFs and a Robotiq \textit{2F-85} parallel jaw gripper with one DOF. A \textit{Structure IO} depth camera is mounted on the wrist of the robot arm (see Figure~\ref{fig:single_viewpoint}). The robot is mounted to the same table on which the objects to be grasped are placed. Our object set is shown in Figure~\ref{fig:robot_objects}. We chose objects for which at least one side fits into the robot hand (preferably more than one). All objects weigh less than 0.5kg.

The computer used for these experiments has an Intel i7-7800X 3.50GHz CPU, 64GB system memory, and an Nvidia GeForce GTX 1080 graphics card with 8GB of memory. All robot experiments were run on ROS
Melodic~\cite{quigley_2009}. For inverse kinematics, an analytic solution was used~\cite{hawkins_gt2013}. Motion planning and collision checking was done in OpenRAVE~\cite{diankov_cmu2008} and trajopt~\cite{schulman_ijrr2014}. We use toppra~\cite{pham_tro2018} to produce a trajectory from a path under joint velocity and acceleration constraints.

To generate grasp poses, the robot moves its hand to a single viewpoint and obtains a point cloud from its wrist-mounted depth camera. The viewpoint is chosen such that the system can observe as much as possible of the scene. Our algorithm then takes the point cloud and produces a set of grasp poses. We  sample $n=500$ points from the point cloud. From the $n \times m$ outputs of ROT ($n$ sampled points, $m$ hand orientations), we select the top-$k$ ($k=20$) scoring orientations for each of the $n$ points. To reduce runtime, we then uniformly subsample 300 out for the $n \times k$ grasp poses and calculate grasp images for those. Next, we select the top-k ($k=150$) scores from GC. 

\begin{figure}
  \begin{center}
    \subfloat[]
    {
      \includegraphics[width=0.1\textwidth]{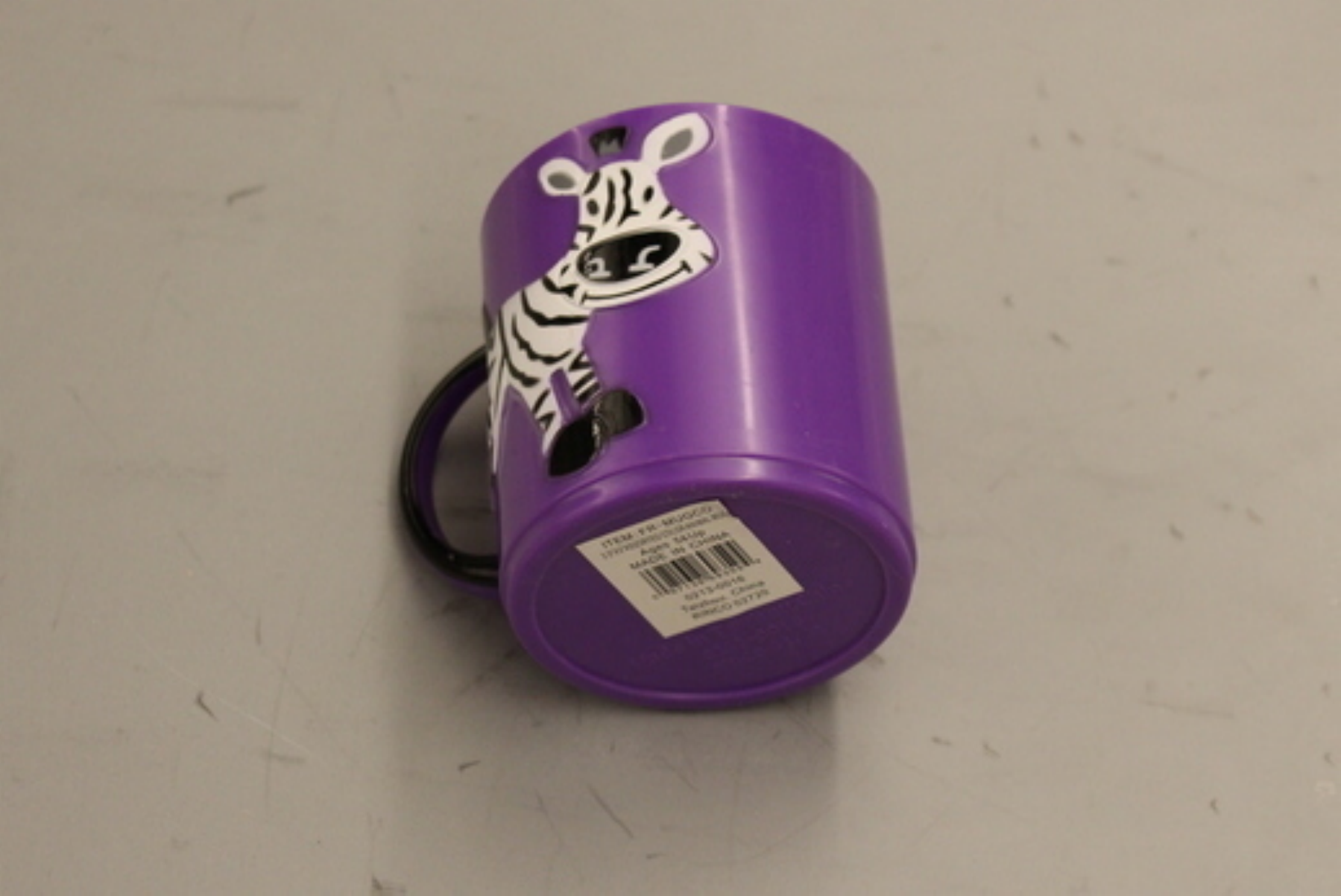}
      \label{fig:mug1}
    }
    \subfloat[]
    {
      \includegraphics[width=0.1\textwidth]{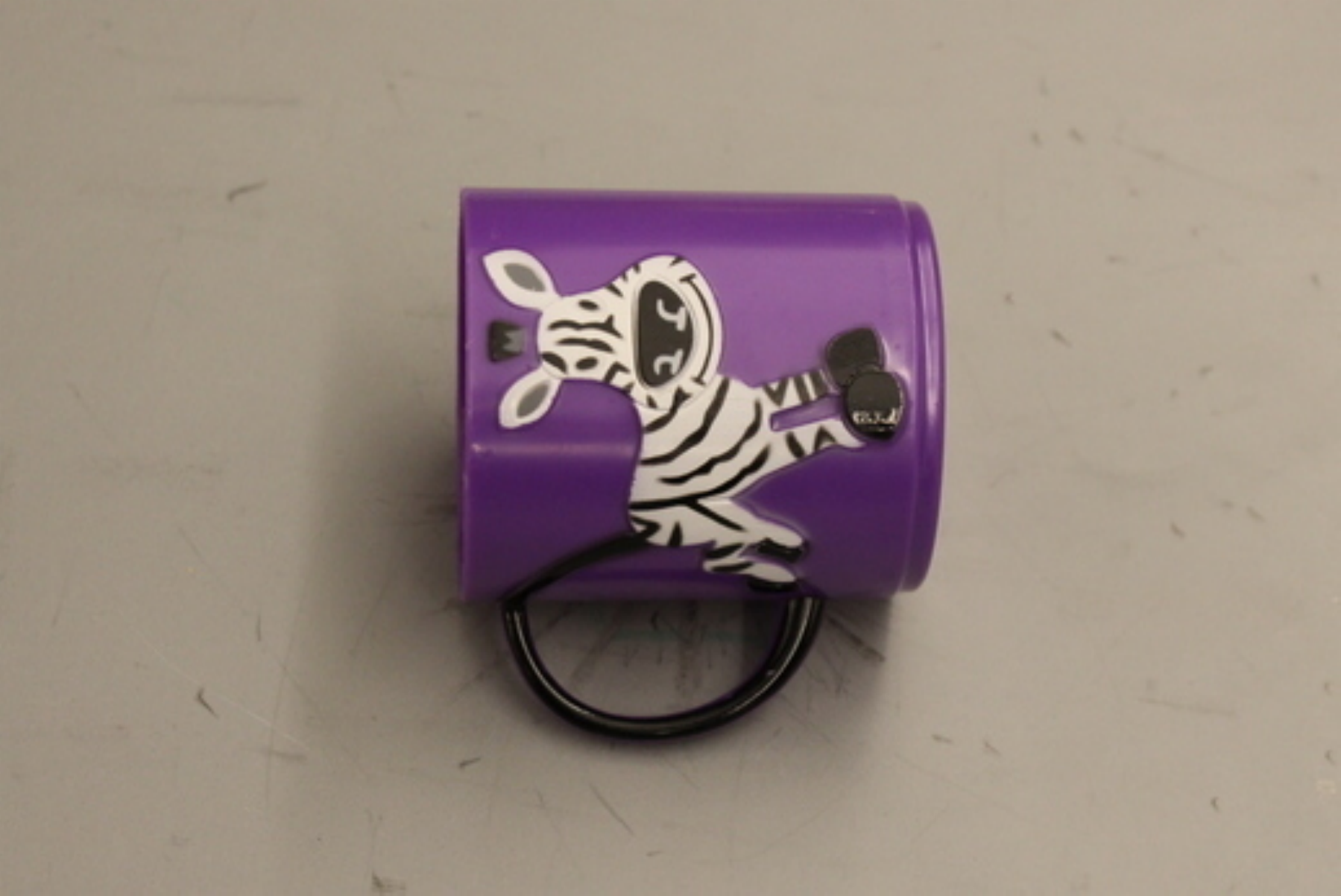}
      \label{fig:mug2}
    }
    \subfloat[]
    {
      \includegraphics[width=0.1\textwidth]{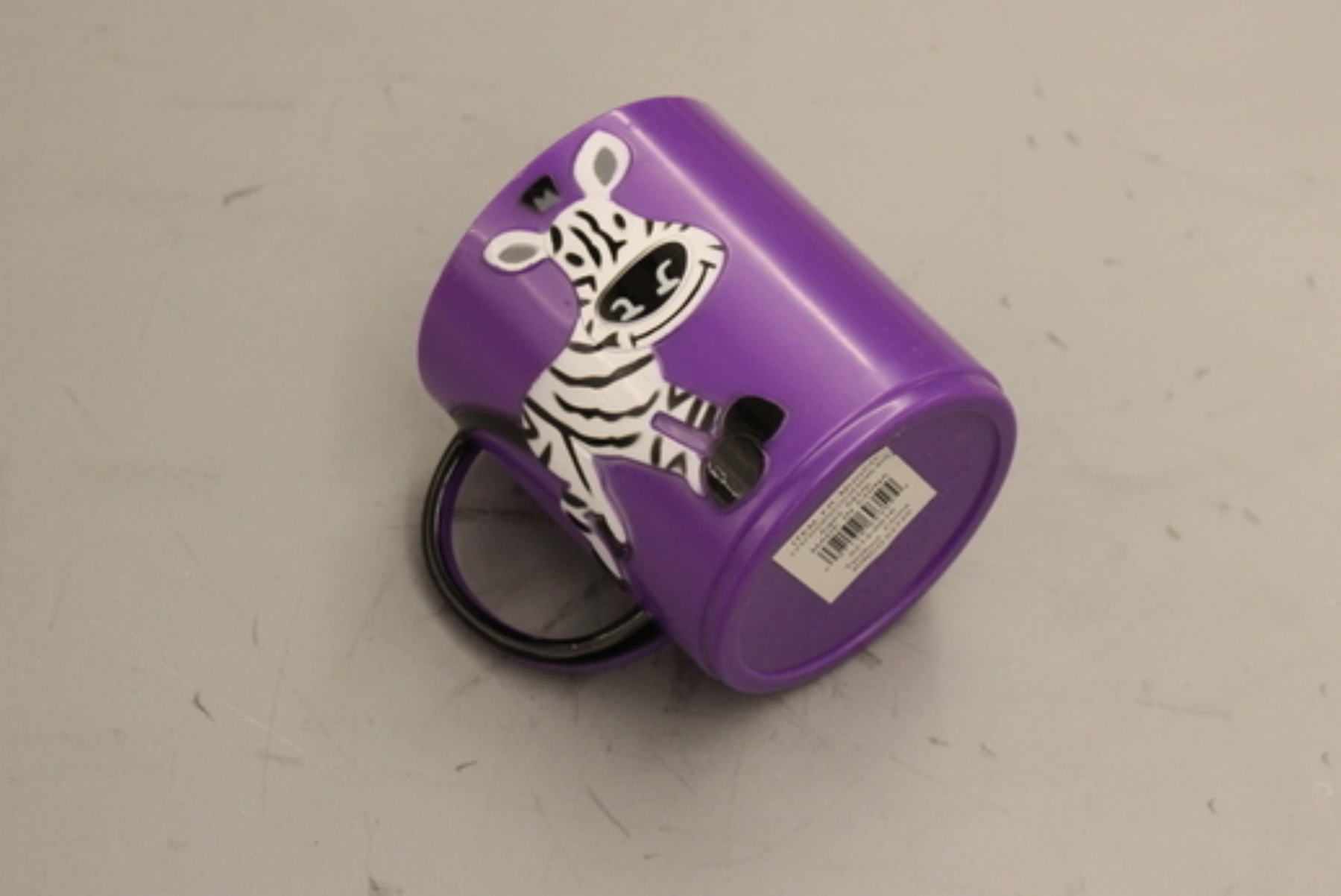}
      \label{fig:mug3}
    }
    \subfloat[]
    {
      \includegraphics[width=0.1\textwidth]{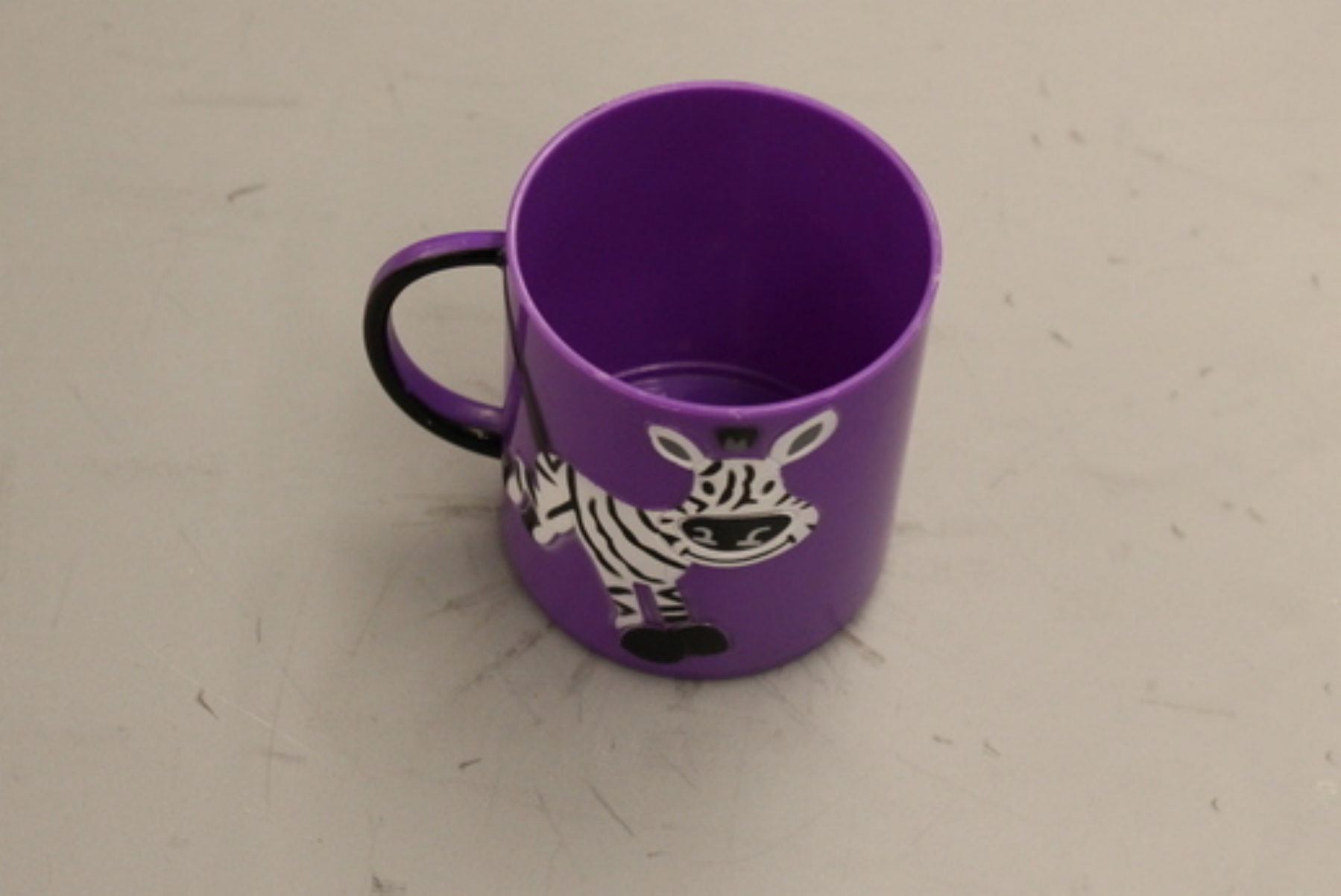}
      \label{fig:mug4}
    }
    \\
    \subfloat[]
    {
      \includegraphics[width=0.1\textwidth]{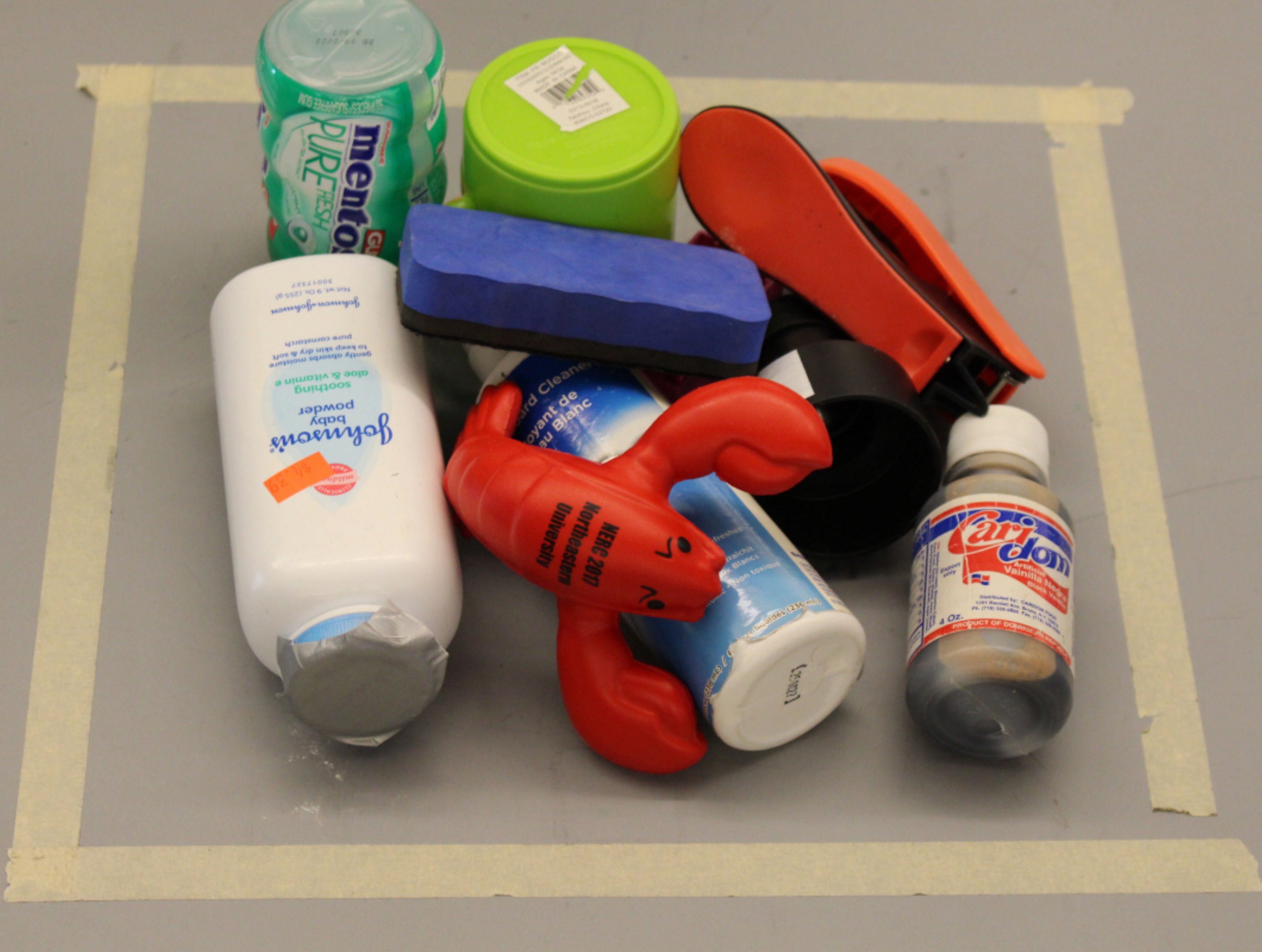}
      \label{fig:clutter1}
    }
    \subfloat[]
    {
      \includegraphics[width=0.1\textwidth]{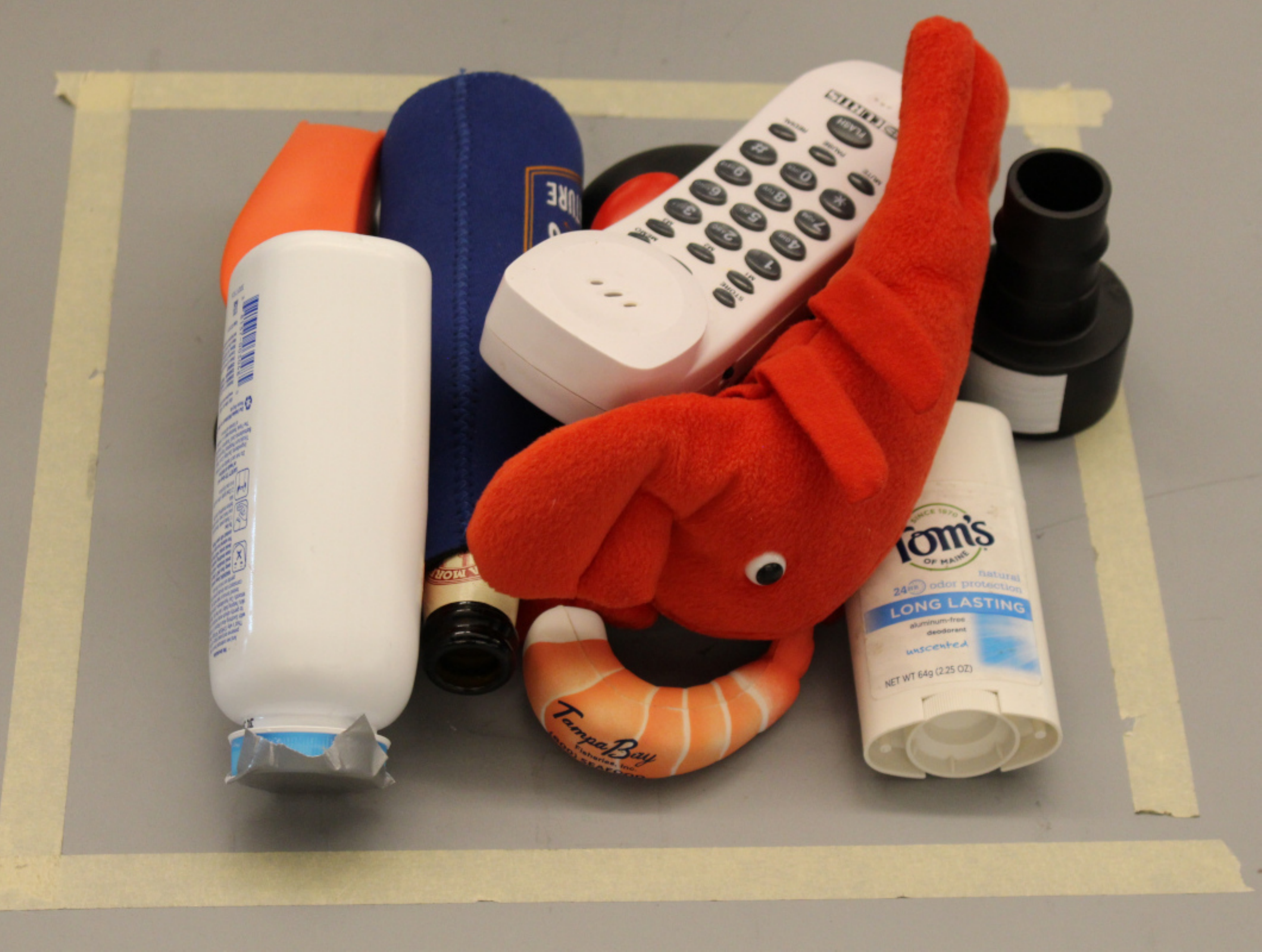}
      \label{fig:clutter2}
    }
    \subfloat[]
    {
      \includegraphics[width=0.1\textwidth]{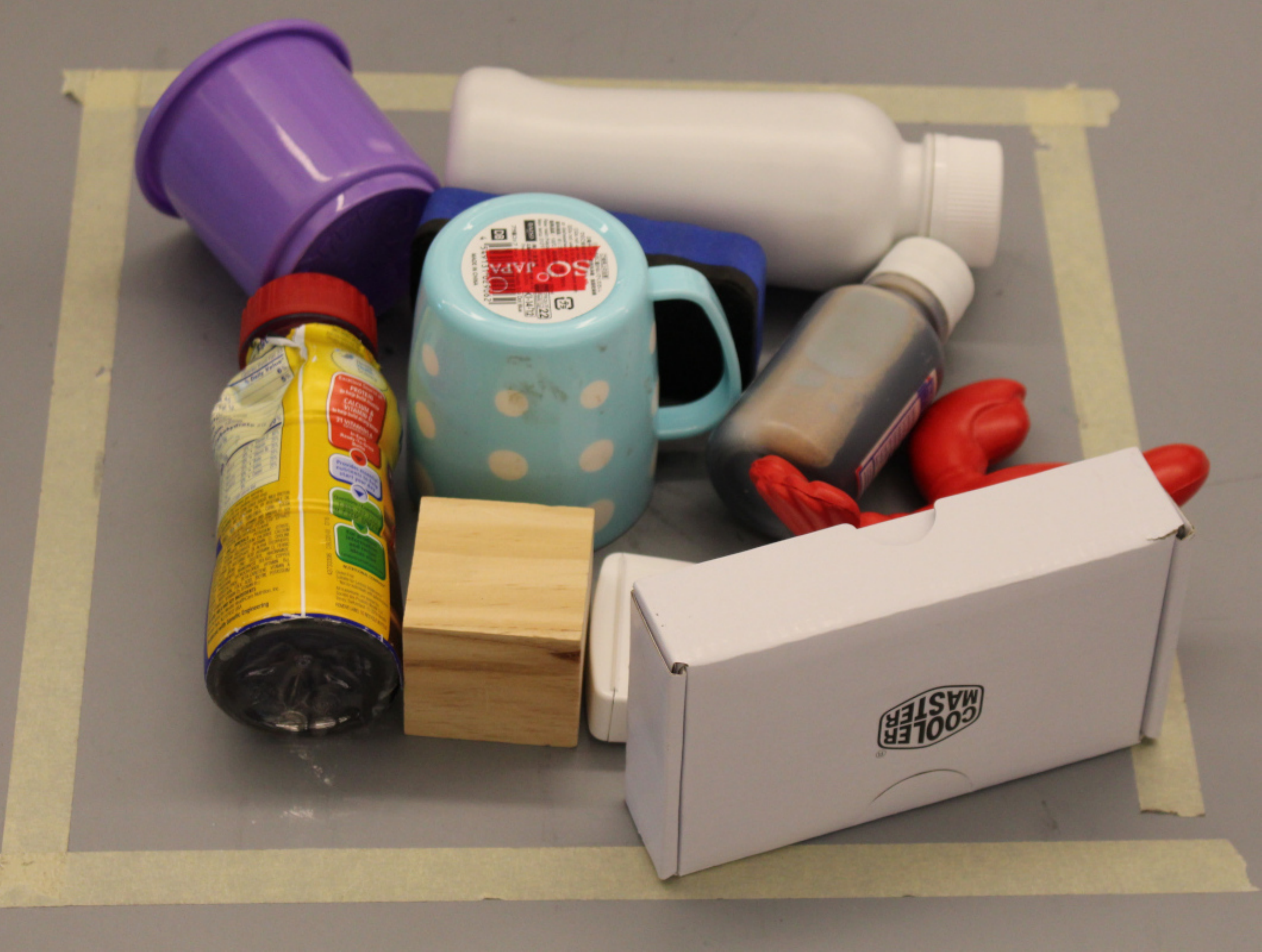}
      \label{fig:clutter3}
    }
    \subfloat[]
    {
      \includegraphics[width=0.1\textwidth]{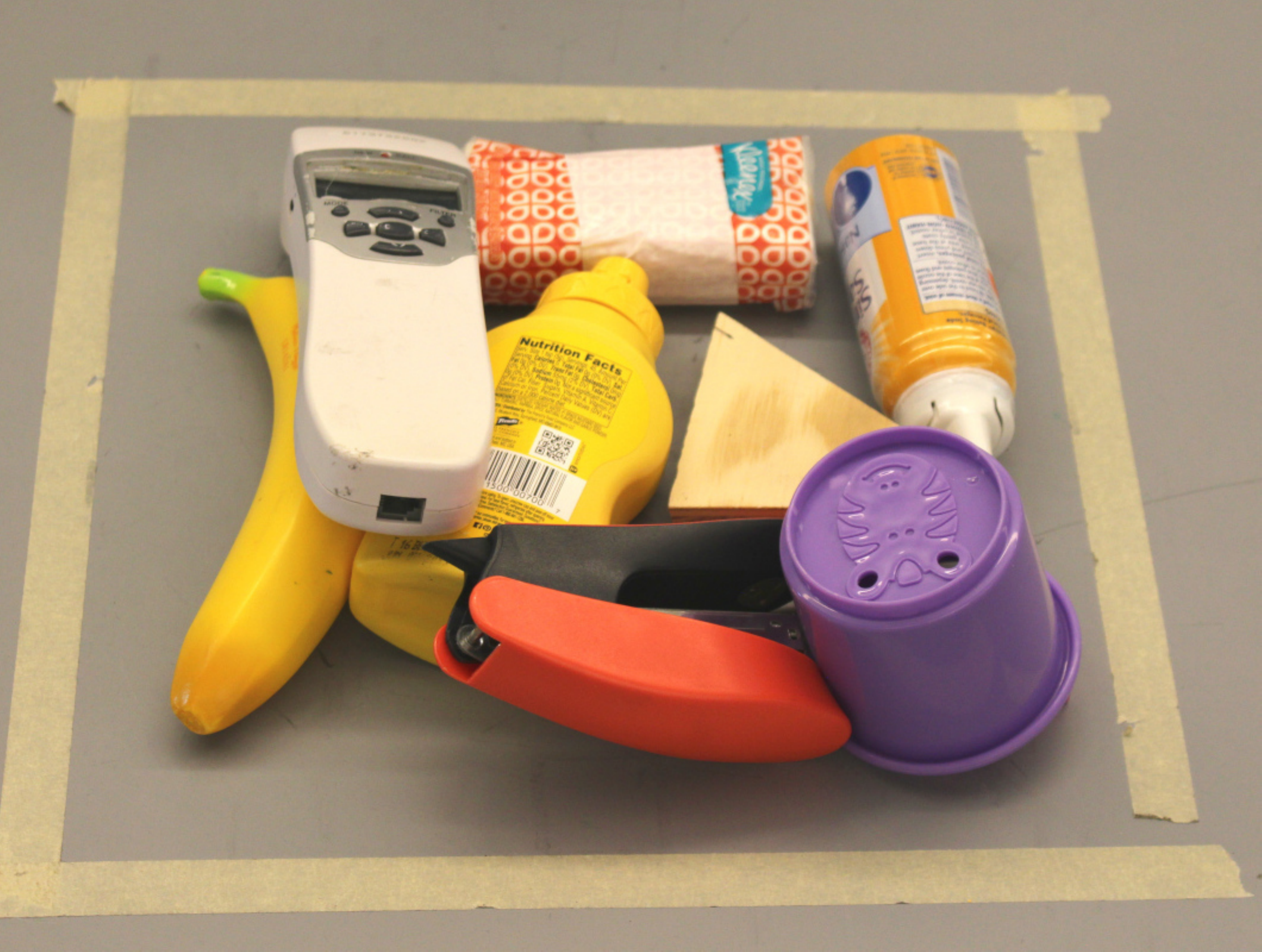}
      \label{fig:clutter4}
    }
  \end{center}
  \caption{Robot experiments setup. (a-d) Examples of an isolated object scene
  with the four different configurations. (e-f) Examples of dense
  clutter scenes.}
  \label{fig:robot_scenes}
\end{figure}

During initial testing, we found that grasps found using our method sometimes generated collisions with the object due to small kinematic errors in robot arm positioning. We believe these errors occur because the training set we used does not take proximity to collision into account. In order to correct for this problem on the robotic system, we used the following procedure. First, we translate each detected grasp pose in the hand closing direction such that the visible points are centered in the grasp. Then, for grasp poses where the hand is further than a threshold distance (half the finger length) from the closest point in the hand closing region, we translate the hand forward along its approach direction as far as possible without generating collisions. Next, we find sets of geometrically aligned grasps using a similar procedure as in~\cite{tenpas_isrr2015} and generate one additional grasp per cluster that corresponds to the geometric center of the cluster. These centers are often close to object (part) centers.  These centers and all grasps produced previously are then checked for collisions with the point cloud. We select a grasp to be executed in a hierarchical manner by considering grasps in the following order: cluster centers, cluster inliers, grasps with a score above 0.5, and grasps with a score below 0.5. For each group, we solve inverse kinematics for all grasps in the group and then compare them using heuristics that take into account the height of the grasp, how vertical the approach direction is and how wide the hand needs to be opened (similar to~\cite{tenpas_ijrr2017}). The former step reduces the number of inverse kinematics problems to be solved, and the second step can improve grasp success on in cluttered scenes because it mitigates the problem of collisions with objects located below other objects in a pile. If no grasp pose can be reached by the robot arm, the next group of grasps is considered.

\subsection{Isolated Object Grasping}

Here, we characterize how well our method is able to grasp objects that are placed in isolation on a table in front of the robot. For each object, we attempt a grasp on up to four distinct configurations:\footnote{The bowls have one configuration that is repeated four times, and the wooden cube has two (a and c).} one to three horizontal orientations and one upright (see Figure~\ref{fig:robot_scenes}(a-d)). We generate a point cloud using a single depth sensor viewpoint and select a grasp using the procedure described in Section~\ref{sect:robot_setup}. Table~\ref{tab:robot_results} (first column) shows the results. The robot achieved an overall grasp success rate of 90.27\% (204 successes out of 226 attempts). Grasps on box corners and the edges of objects were the most common failure modes in this experiment.

\subsection{Dense Clutter Grasping}

Here, we evaluate our method in a dense clutter setting similar to the one described in~\cite{tenpas_ijrr2017}. First, we place ten randomly selected objects from the object set into a box. The object set for this experiment is the same as shown in Figure~\ref{fig:robot_objects}, except for the three bowls. We then turn the box upside down on a table in a fixed location, shake the box, and remove it. Finally, the robot tries to grasp the objects one by one until no objects remain on the table. Examples of clutter scenes are depicted in Figure~\ref{fig:robot_scenes}(e-h).

Results are shown in the second column of Table~\ref{tab:robot_results}. Overall, the robot attempted 176 grasps out of which 144 were successful (81.82\% success rate). The most common failure modes were edge grasps and corner grasps on boxes some of which may have failed due to noise in the robot's calibration. Furthermore, the robot removed 144 out of 150 objects from the table (96\% success rate). The reason for the six objects which were not removed is that they rolled out of the robot's reach or view because of collisions which occurred while grasping another object.

\begin{table}
  \caption{Results of robot grasping experiments.}
  \label{tab:robot_results}
  \begin{center}
    \begin{tabular}{c|c|c}
     & Isolated objects & Dense clutter \\ \hline
     Num grasp attempts & 226  & 176 \\ \hline
     Num grasp successes & 204 & 144  \\ \hline
     Grasp success rate & 90.27\%  & 81.82\% \\ \hline
     Object removal rate & / & 96.00\% \\
    \end{tabular}
  \end{center}
\end{table}


\section{Conclusion}
\label{sec:conclusion}
We proposed a method for predicting grasp poses which were represented as height maps of an orthographically projected point cloud. In simulation experiments, we found that our method is both fast and precise, in particular compared to geometric grasp proposal generation. We also showed that we can learn grasps based on a physics simulation. Finally, we demonstrated that our method can effectively grasp objects in isolation and in dense clutter on a robot.

In future work, we could extend our method to learn collisions with an object's support plane, like a table, and with its surroundings, like in cluttered scenes. Another interesting direction is to learn task-dependent grasps, e.g., to detect grasps on the handle of a drill. Typically, task dependency is learned separately from grasp pose detection but could be learned together with grasp quality.






\section*{ACKNOWLEDGMENT}
We thank Andrea Baisero, Marcus Gualtieri, and David Klee for reviewing drafts
of this paper and the anonymous reviewers for their insightful comments.
Funding was provided by NSF 1724257, NSF 1724191, NSF 1763878, NSF 1750649, and
NASA 80NSSC19K1474.


\bibliographystyle{IEEEtran}
\bibliography{IEEEabrv,refs}

\end{document}